\documentclass[runningheads]{llncs}
\usepackage{pifont}
\usepackage[T1]{fontenc}
\usepackage{graphicx}
\usepackage{hyperref}
\usepackage{color}
\usepackage[table]{xcolor}
\usepackage{amsmath}
\usepackage{placeins}
\usepackage{float}  

\usepackage{lipsum}  % Optional: For dummy text
\usepackage{array}

\begin{document}
\title{PosePilot: An Edge-AI Solution for Posture Correction in Physical Exercises}

\newcommand{\repeatthanks}{\textsuperscript{\thefootnote}}
\author{Rushiraj Gadhvi\inst{1,2}\orcidID{0009-0007-7976-5506}\thanks{These authors contributed equally to this work} \and
Priyansh Desai\inst{1,3}\orcidID{0009-0009-8769-6912}\repeatthanks \and
Siddharth\inst{1,4}\orcidID{0000-0002-1001-8218}}

\authorrunning{R. Gadhvi, P. Desai, Siddharth et al.}
\titlerunning{PosePilot: An Edge-AI Posture Correction Solution}
\institute{HTI Lab, Plaksha University, Mohali, India \and
\email{rushiraj.gadhvi@plaksha.edu.in} \and
\email{priyansh.desai@plaksha.edu.in} \and
\email{siddharth.s@plaksha.edu.in}}

\maketitle
\begin{abstract}

Automated pose correction remains a significant challenge in AI-driven fitness systems, despite extensive research in activity recognition. This work presents PosePilot, a novel system that integrates pose recognition with real-time personalized corrective feedback, overcoming the limitations of traditional fitness solutions. Using Yoga, a discipline requiring precise spatio-temporal alignment as a case study, we demonstrate PosePilot’s ability to analyze complex physical movements. Designed for deployment on edge devices, PosePilot can be extended to various at-home and outdoor exercises. We employ a Vanilla LSTM, allowing the system to capture temporal dependencies for pose recognition. Additionally, a BiLSTM with multi-head Attention enhances the model’s ability to process motion contexts, selectively focusing on key limb angles for accurate error detection while maintaining computational efficiency. As part of this work, we introduce a high-quality video dataset used for evaluating our models. Most importantly, PosePilot provides instant corrective feedback at every stage of a movement, ensuring precise posture adjustments throughout the exercise routine. The proposed approach 1) performs automatic human posture recognition, 2) provides personalized posture correction feedback at each instant which is crucial in Yoga, and 3) offers a lightweight and robust posture correction model feasible for deploying on edge devices in real-world environments. 
\end{abstract}

\section{Introduction}
Correcting human postures is crucial across diverse physical activities spanning rehabilitation, Yoga, and physiotherapy. Posture correction is also utilized in sports training to refine follow-throughs and ensure proper posture during actions like hitting a ball \cite{Lees,NRBM,Bolling,kelly}. Literature also proposes systems to analyze and improve postures in sports like basketball \cite{pai}, badminton \cite{shan,lin}, and tennis \cite{vives}. Similarly, physiotherapy and rehabilitation regimens frequently require accurate adjustment of postures to achieve the best results. For instance, Raza et al. presented a systematic approach for correcting postures to improve fitness \cite{Raza}, while Taati et al. \cite{taati} developed an assessment system to detect and categorize compensation during robotic rehabilitation therapy. 

While fitness routines offer substantial benefits for individual health and well-being, incorrect execution may potentially lead to injuries \cite{bianchi,che}. Unfortunately, existing systems lack personalized posture correction, relying on predefined correct/incorrect decision parameters set by domain experts, that may not represent practitioners' diversity including age, weight, height, flexibility level, etc. Recent advances in artificial intelligence (AI) may support the development of systems capable of providing personalized pose correction feedback.

Across various physical exercises, we specifically chose Yoga to build and evaluate our AI model for several key reasons. In the last couple of decades, Yoga has regained popularity as a fitness routine due to its benefits to both physical and mental health \cite{telles}. However, Yoga is often perceived merely as achieving a set of final postures. This is misleading since Yoga's true benefits manifest when the correct sequence of sub-poses is performed leading to the final pose. This not only enhances the efficacy of Yoga but also reduces the risk of injury. Posture correction being crucial in Yoga motivated us to develop a generalized posture recognition and correction system.

Deploying AI models on edge devices improves accessibility and reliability by eliminating internet dependence and cloud transmission. Additionally, local computation enhances privacy by keeping sensitive data on the device. This ensures a secure, reliable, and independent fitness experience, making such solutions more practical. %Real-time posture correction may benefit from long-short term memory (LSTM)-based models, making them ideal for low-latency, resource-constrained environments. 
Research has shown that LSTM-based architectures can process sequential data significantly faster than Transformers, making them suitable for low-latency applications on edge devices \cite{Gad}.

% this para can be reduced if needed
In this work, we present PosePilot\footnote{\href{https://github.com/gadhvirushiraj/PosePilot}{https://github.com/gadhvirushiraj/PosePilot}}, an innovative system for correcting human posture in physical activities evaluated on Yoga postures (Figure \ref{fig:pipeline}). While most literature has focused only on pose recognition, PosePilot also encompasses the capacity to identify errors while performing a Yoga \textit{asana} (pose) and providing personalized correction feedback to the user. This is done by utilizing a Bidirectional Long Short-Term Memory (BiLSTM) with multi-head Attention to model the temporal sequential nature of all joints' movements from the inception to the culmination for each \textit{asana}. Thus, PosePilot provides a comprehensive fitness solution that can perform both pose recognition and provide personalized correction feedback based on detected errors in the pose at every time interval. %while being optimal for resource-constrained environments.
The choice of models ensures that the system remains computationally efficient for deployment on edge devices. 

\begin{table}[h]
\centering
\caption{Comparison of previous studies on pose recognition and correction capabilities (PR: Pose Recognition; PC: Pose Correction; ED: Edge Deployability)}
\label{tab:my_table}
\setlength{\tabcolsep}{3pt}
\begin{tabular}{|>{\raggedright}m{2cm}|>{\raggedright}m{1.7cm}|>{\raggedright}m{2.4cm}|>{\centering\arraybackslash}m{1.9cm}|>{\centering\arraybackslash}m{1.4cm}|>{\centering\arraybackslash}m{1.4cm}|}
\hline
\rowcolor{gray!20} % Light gray background for header
\parbox[c]{2cm}{\centering\rule{0pt}{2.8ex}\textbf{Study}} & 
\parbox[c]{1.7cm}{\centering\rule{0pt}{2.7ex}\textbf{Dataset}} & 
\parbox[c]{2.4cm}{\centering\rule{0pt}{2.8ex}\textbf{Model Overview}} & 
\parbox[c]{1.9cm}{\centering\rule{0pt}{2.8ex}\textbf{PR Accuracy}} & 
\parbox[c]{1.4cm}{\centering\rule{0pt}{2.8ex}\textbf{PC}} & 
\parbox[c]{1.4cm}{\centering\rule{0pt}{2.8ex}\textbf{ED}} \\\hline
Yadav et al. \cite{Yadav19} & Yadav et al. \cite{Yadav19} & OpenPose + CNN + LSTM & 98.92\% & \ding{55} & \ding{55} \\
\hline
Infinity Yoga Tutor \cite{Infinity_Yoga_Tutor} & Yadav et al. \cite{Yadav19} & OpenPose + CNN + LSTM & 99.91\% & \ding{55} & \ding{55} \\
\hline
Pose Tutor \cite{Pose_Tutor} & Yoga-82 \cite{Yoga82}, Pilates-32, Kungfu-7  & Densenet + KNN & 79\%, 82\%, 81\% & \ding{51}  & \ding{55} \\
\hline
Yog-Guru \cite{yog-guru} & Yoga-82 \cite{Yoga82}  & OpenPose + CNN & 95\% & \ding{55}  & \ding{55}  \\
\hline
YoNet \cite{YoNet} & Yoga-82 \cite{Yoga82} & YoNet & 94.91\% & \ding{55}  & \ding{55}  \\
\hline
3D-CNN \cite{Jain} & 3D-CNN \cite{Jain} & 3D CNN & 91.15\% & \ding{55} & \ding{55}  \\
\hline
\textbf{PosePilot (This work)} & \textbf{In-house} & \textbf{LSTM(PR) \& BiLSTM + Multi-Head Attention(PC)} & \textbf{97.52\%} & \textbf{\ding{51}} & \ding{51} \\
\hline
\end{tabular}
\end{table}

\section{Literature Survey}

\subsection{Pose Recognition}
Researchers have explored wearables, good at protecting privacy, for pose recognition \cite{wang,Novel-IOT}. By attaching accelerometers to the user’s wrist and right hip, Wang et al. achieved an accuracy of 91.5\% for recognizing standing, sitting, and walking poses. However, carrying around sensors and a battery pack could be impractical for long durations. In vision-based systems, the majority of the research has primarily focused on classifying the final pose. Certain Yoga poses, such as \textit{Sukhasana} and \textit{Padmasana}, closely resemble each other in terms of their final pose. For such poses, vision-based systems face challenges in classifying the pose from just the final video frame due to variations in camera viewpoints. However, leveraging temporal information leading up to the pose could provide valuable information to improve recognition.

Towards modeling this temporal dynamics, Yadav et al. \cite{Yadav19} proposed a pipeline for Yoga pose detection, utilizing a hybrid architecture combining a convolution neural network (CNN) for feature extraction and an LSTM network for sequence analysis. This approach achieved a promising 99.38\% accuracy across six \textit{asanas} using 45-frame videos. Similarly, the Infinity Yoga Tutor system employed a method that first extracted key points, followed by a time-distributed 1D CNN and LSTM, achieving 99.91\% accuracy \cite{Infinity_Yoga_Tutor}. Jain et al. \cite{Jain} applied a 3D CNN model to classify twelve Yoga poses, obtaining 91.5\% accuracy on their dataset and 99.39\% on Yadav et al.'s dataset \cite{Yadav19}. Thus, we discovered that calculating joint angles could alone provide sufficient information for pose recognition, and also remove additional overhead of feature extraction via the CNN layers. This is crucial while designing a system capable of working in real-time on an edge device in diverse real-world settings for continuous feedback.

\subsection{Pose Correction}
Few studies have focused on pose correction, particularly for Yoga (Table \ref{tab:my_table}). Thar et al. \cite{Thar} proposed a method that extracts keypoints through pose estimation, calculates joint angles, and compares them to an instructor's pose. The system computes deviations in pose angles using an average angle error threshold across four ranking levels. Gradient-based skeleton visualization is used to indicate correction angles for joints. Pose Tutor \cite{Pose_Tutor} does corrections by analyzing the pose's joint probability distributions and evaluating the likelihood of deviations to detect incorrect angles. Recent studies in ergonomic settings demonstrate real-time feedback efficacy. Kim et al. \cite{Kim2023} developed a posture correction system using IMU sensors, triggering alerts when neck flexion exceeded 25 degrees or trunk slouching surpassed thresholds, reducing muscle strain by 27\% and improving alignment. While these solutions can only identify incorrect angles based on pre-set thresholds, they lack a precise mechanism to suggest corrections. Furthermore, these solutions overlook inaccuracies in sub-poses throughout the sequence making it impossible to provide personalized feedback. 

Dittakavi et al. \cite{Dittakavi2024CARE} developed the CARE system, which uses counterfactual explanations following pose classification to identify adjustments to angles that could lead to a different outcome. It then applies algorithmic recourse to select the best option among the counterfactuals generated. In the final step, the system presents an action vector, outlining the necessary movements for the user to achieve proper posture. A key limitation of the system is that it only provides feedback on the final pose, without addressing corrections for the intermediate stages (sub-poses) of the movement. We aim to address this issue so that users can get actionable feedback for the whole temporal sequence which is crucial for the user to correctly perform Yoga \textit{asanas}.

\section{Methodology}

\subsection{In-house Dataset} 
\label{sec:inhouse-dataset}
\begin{figure}[H]
    \centering
    \includegraphics[width=0.6\textwidth]{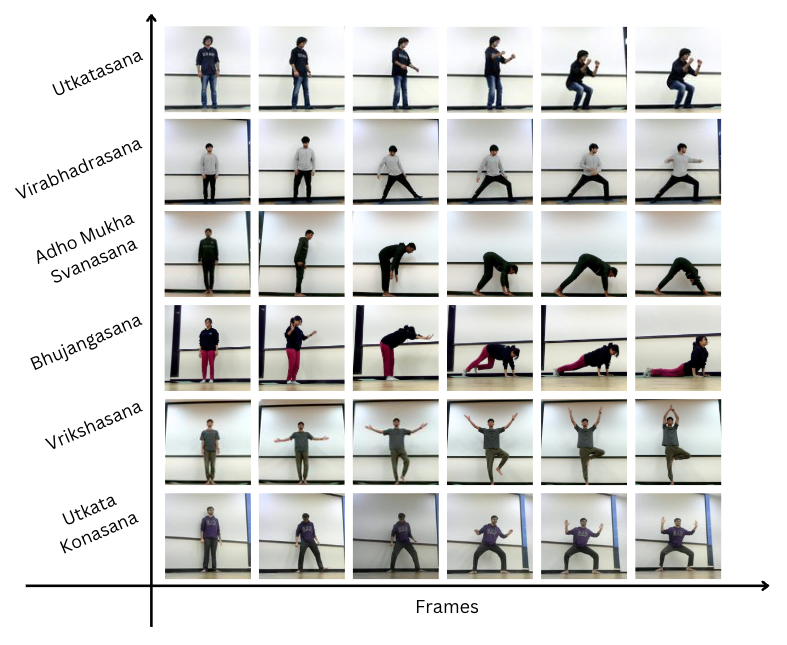}
    \caption{Sample of In-house Dataset.}
    \label{fig:dataset-sample}
\end{figure}

To date, there is no high-quality publicly available video-based benchmarking dataset for Yoga \textit{asana}, although several image datasets exist that capture only final static poses like Yoga-82 \cite{Yoga82}. While Yadav et al. \cite{Yadav19} introduced a video-based dataset, it is limited by single-angle capture, suboptimal video quality and lighting conditions, a small number of recordings per pose, and low participant diversity. Additionally, Yoga practitioners in these datasets are generally not professionals, thus limiting the establishment of the ground truth baseline for pose correction. To develop a more robust solution, a new dataset was created capturing each pose from four angles. Fourteen participants (10 male and 4 female, age range 17-25 years) with varied body structures were recorded performing six Yoga \textit{asanas}: \textit{Vrikshasana} (Tree Pose), \textit{Utkatasana} (Chair Pose), \textit{Virabhadrasana} (Warrior Pose), \textit{Bhujangasana} (Cobra Pose), \textit{Adho Mukha Svanasana} (Downward Dog), and \textit{Utkata Konasana} (Goddess Pose)(Figure \ref{fig:dataset-sample}). Filming was conducted indoors with controlled lighting, primarily with bright white light. The final dataset consisted of 56 videos per pose, each with an average duration of 7 seconds, totaling 336 videos.

\begin{equation}
    \Theta = \arccos\left(\frac{\mathbf{x} \cdot \mathbf{y}}{\|\mathbf{x}\| \|\mathbf{y}\|}\right)
    \label{eq:4}
\end{equation}

Mediapipe \cite{Mediapipe} was utilized to analyze the video footage frame-by-frame for keypoint extraction. Each frame yielded 33 distinct keypoints. Subsequently, to facilitate the calculation of angles for pose analysis, 16 keypoints were excluded based on relevance and domain knowledge. Using the remaining 17 keypoints, a total of 680 angles were computed (Equation \ref{eq:4}) based on all combinations of three keypoints. Here, \(\mathbf{x}\) and \(\mathbf{y}\) denote vectors originating from the central keypoint (the vertex of the angle) to its two adjacent keypoints. These angles served as features for the machine learning models, with the dataset split in an 80:20 ratio for training and testing. The final dataset will be made publicly available within the code repository.

\subsection{Pose Recognition Model}
Our goal is to build a model that can classify Yoga \textit{asanas} using sequential images in the video as input. However, since each pose sequence is recorded at practitioner's own pace, the video sequences vary in length from person to person and contain a substantial amount of redundant information. To address this, we extract the \textit{key frames}, a subset of frames that capture the most significant changes in body posture while performing pose. To do so, we calculated the aggregated standard deviation across all 680 angles for each frame at time $t$ with respect to the frames ranging in the interval $t - 2$ to $t + 2$. (Equations \ref{eq:1}-\ref{eq:3})

% \noindent
% \begin{minipage}{\linewidth}
% \begin{equation}
% \label{eq:1}
% \mu_{tj} = \frac{1}{p} \sum_{i = -\lfloor \frac{p}{2} \rfloor}^{\lfloor \frac{p}{2} \rfloor} F_{t+i,j} \quad \text{(where $p =$ interval length)}
% \end{equation}

% \vspace{-1em}
% \begin{equation}
% \label{eq:2}
% \sigma_{tj} = \sqrt{\frac{1}{p-1} \sum_{i = -\lfloor \frac{p}{2} \rfloor}^{\lfloor \frac{p}{2} \rfloor} (F_{t+i,j} - \mu_{tj})^2}
% \end{equation}

% \vspace{-1em}
% \begin{equation}
% \label{eq:3}
% E(\sigma_{t}) = \frac{1}{q} \sum_{j=1}^{q} \sigma_{tj} \quad \text{(where $q =$ number of angles)}
% \end{equation}

% \vspace{0.5em}
% \captionsetup{type=equation, font=small, labelfont=it, textfont=it}
% \caption{Let \(F_{i,j}\) denote the value of the \(j\)th angle (out of 680) at frame \(i\). Using this, the local mean \(\mu_{tj}\) and standard deviation \(\sigma_{tj}\) are computed for each angle \(j\) within a temporal window of length \(p\) centered at frame \(t\). The aggregated deviation \(E(\sigma_t)\) is then calculated by averaging the standard deviations across all \(q\) angles at frame \(t\), providing a measure of overall posture variability over time.}
% \end{minipage}

\begin{subequations}
\label{eq:all}
\begin{align}
\mu_{tj} &= \frac{1}{p} \sum_{i = -\lfloor \frac{p}{2} \rfloor}^{\lfloor \frac{p}{2} \rfloor} F_{t+i,j} 
\quad\text{(where \( p \) = interval length)} \label{eq:1} \\
\sigma_{tj} &= \sqrt{\frac{1}{p-1} \sum_{i = -\lfloor \frac{p}{2} \rfloor}^{\lfloor \frac{p}{2} \rfloor} (F_{t+i,j} - \mu_{tj})^2}
\label{eq:2} \\
E(\sigma_{t}) &= \frac{1}{q} \sum_{j=1}^{q} \sigma_{tj}
\quad\text{(where \( q \) = number of angles)} \label{eq:3}
\end{align}
\end{subequations}

\vspace{0.5em}
{\small\textit{Equation~\ref{eq:all}: \(F_{i,j}\) denotes the value of the \(j\)th angle (out of 680) at frame (\(i\)). Using this, the local mean \(\mu_{tj}\) and standard deviation \(\sigma_{tj}\) are computed for each angle (\(j\)) within the temporal window of length (\(p\)) centered at frame (\(t\)). The aggregated deviation \(E(\sigma_t)\) is then calculated by averaging the standard deviations across all (\(q\)) angles at frame (\(t\)), providing a measure of overall posture variability over time.}}

We then computed the local maxima across the aggregated average standard deviations $E(\sigma_t)$ to identify the frames with the highest deviations, signaling significant posture movement. Our hypothesis was that such key frames would have critical information for our pose recognition model. The number of \textit{key frames} with the greatest deviations ($k$) serves as a hyperparameter in our approach. During testing, we found that $k = 10$ yielded optimal results with typically short pose sequences. Furthermore, $k$ can also be adjusted based on the length of the pose sequences and the level of similarity between poses in the dataset.

\begin{figure}[h]
    \centering
    \includegraphics[width=\textwidth]{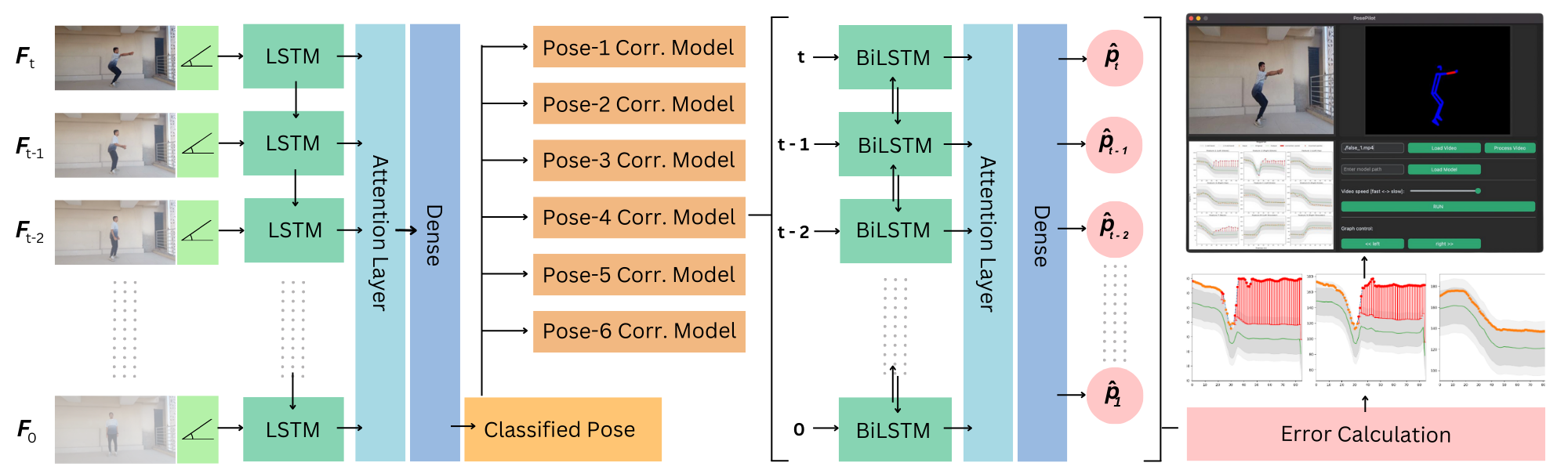}
    \caption{PosePilot Overview. Video frame's extracted joint angles from time $[0,t]$ are first fed into an LSTM that captures motion and classifies the current yoga pose, selecting one of six trained correction models. A BiLSTM then analyzes those frames to forecast the next joint-angle vector $\hat p_{t}$. These deviations are processed by an error-calculation module, and the user interface presents per-angle correction feedback alongside a dynamic visualization of pose improvement.}
    \label{fig:pipeline}
\end{figure}

Data augmentation was performed by introducing noise to the respective angles according to their distributions within the sequence. This augmentation enhances model robustness by exposing it to slight angle variations, helping it detect correctly the intended \textit{asana} even when user performs it incorrectly. However, this approach must be applied carefully to avoid generating angle values that fall outside the range of human possibility. For classification, we employed a single‑layer LSTM with multi‑head attention, followed by a feed‑forward neural layer: at each time step, the input of the LSTM was the 680-dimensional vector of joint angles for the \textit{key frames} identified, produced a probability distribution over the six asanas, from which the highest scoring class was chosen (see Figure \ref{fig:pipeline}). This choice was made due to the LSTM's ability to handle sequential data, making it ideal for analyzing temporal patterns in physical activity.

\subsection{Pose Correction Model}
The pose correction model utilized the distinct temporal patterns across different angles associated with each pose. We first calculated the average frame count for each pose and standardized the clips to match it. The aggregated standard deviation for each frame at time $ t $, relative to the frames from $ t - 2 $ to $ t + 2 $ across all 680 angles. For sequences with more than the average number of frames, frames with the least standard deviation were removed iteratively to match the average frame count. Conversely, for sequences with fewer frames, we identify the frame at time $ t $ with the highest standard deviation and perform spline interpolation over the window $[t-5, t+5]$ to insert a new frame at $ t + 1 $. This iterative process continues until the sequence reaches the target average frame count, ensuring that each additional frame is placed where it could maximally reduce temporal gaps.
Based on domain knowledge and the inputs by Yoga experts, we reduced the feature size from 680 angles to 9 angles for correction analysis: left shoulder, right shoulder, left elbow, right elbow, left hip, right hip, left knee, right knee, and neck. For the correction model, we avoided noise-based augmentation to preserve exact joint angles, important for precise angle forecasting. Instead, we augmented by shifting each selected frame’s index by $\pm1$  and the corresponding data points were extracted and combined to form the augmented dataset. This ensured robust data augmentation by leveraging temporal offsets within each sequence, enhancing model generalization while preventing unrealistic angles arising from noise-based augmentation techniques.

For our model, we opted for a two-layered BiLSTM network to account for contextual data from both preceding and subsequent states within the sequence. In this way, the model gains richer temporal context, allowing it to effectively capture the general trends of different limb-angles, allowing us to forecast future movements for the practitioner's ongoing pose. During the pose evaluation phase, we conduct a comparison between the model's predictions and the user's movements across the nine angles. If the difference between the predicted and actual angle values exceeds 1.5 standard deviations, the point is flagged as an error, indicating a potential deviation from expected performance requiring correction.

\begin{table}[H]
\centering
\caption{Pose Recognition and Correction Results.}\label{tab:classification_report}
\begin{tabular}{|l|c|c|}
\rowcolor{gray!30}
\hline
\textbf{Activity} & \textbf{Recognition Accuracy(\%)} & \textbf{Correction Model MSE} \\
\hline
\textit{Bhujangasana} & 94.11 & 0.0010 \\
\hline
\textit{Vrikshasana} & 98.00 & 0.0015 \\
\hline
\textit{Utkata Konasana} & 94.44 & 0.0013 \\
\hline
\textit{Utkatasana} & 99.4 & 0.0013 \\
\hline
\textit{Adho Mukha Svanasana} & 100.00 & 0.0018 \\
\hline
\textit{Virabhadrasana} & 99.20 & 0.0014 \\
\hline
\textbf{Average} & \textbf{97.52} & \textbf{0.00138} \\
\hline
\end{tabular}
\end{table}

\section{Evaluation}
\subsection{Pose Recognition}
To ensure the robustness of our pose recognition model, we employed a 10-fold cross-validation approach. This method allowed us to assess the model’s performance across different subsets of data, reducing the likelihood of over-fitting and making the model generalizable. We further evaluated our model by testing its performance on dataset provided by Yadav et al. \cite{Yadav19} and achieved comparable results. In addition, we applied the model to our in-house dataset, which consists of annotated video sequences for six distinct \textit{asanas}. On this dataset, the model achieved an impressive F1 score of 0.99, highlighting its high accuracy across all the different \textit{asanas} as shown in Table \ref{tab:classification_report}. The overall classification accuracy was 97.52\%, correctly identifying 946 out of 970 frames. The results demonstrated that the model is not only robust but also capable of detecting the intended poses even when errors or variations are introduced by the users during performance, such as joint angle deviations in posture
or different execution timing.

\begin{figure}[h]
    \centering
    \includegraphics[width=\textwidth]{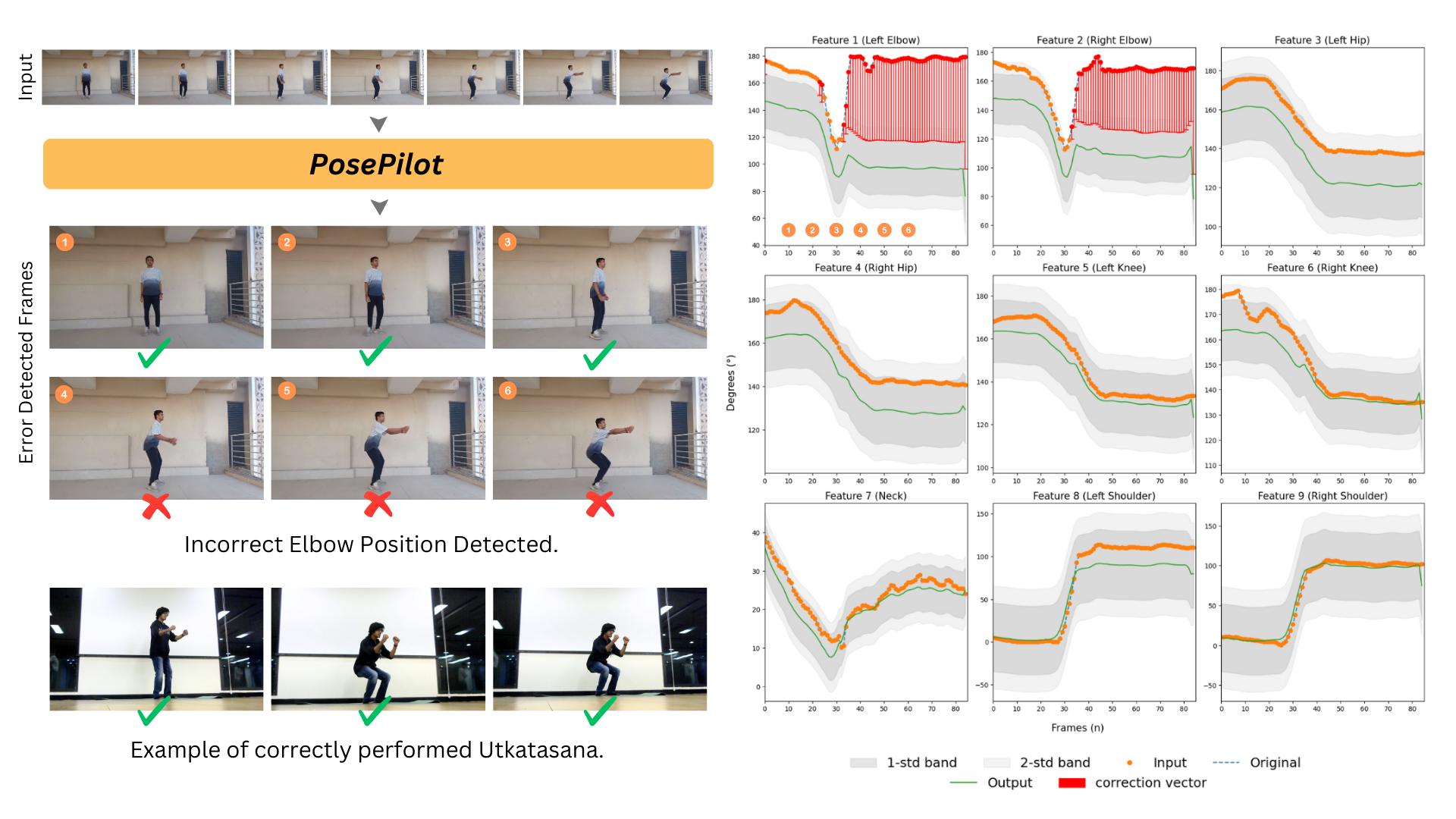}
    \caption{Correction graph for an incorrectly performed \textit{Utkatasana}. Error corresponding frames are marked with red crosses. Any joint angle deviating more than 1.5 standard deviations from the ideal pose is flagged, and red vectors show the adjustment needed to bring each point back within the acceptable range. A correctly performed \textit{Utkatasana} is shown alongside for comparison.}
    \label{fig: Correction Graph}
\end{figure}

\subsection{Pose Correction}
\label{sec:pose_correction}
\begin{figure}[h]
    \centering
    \includegraphics[width=0.75\textwidth]{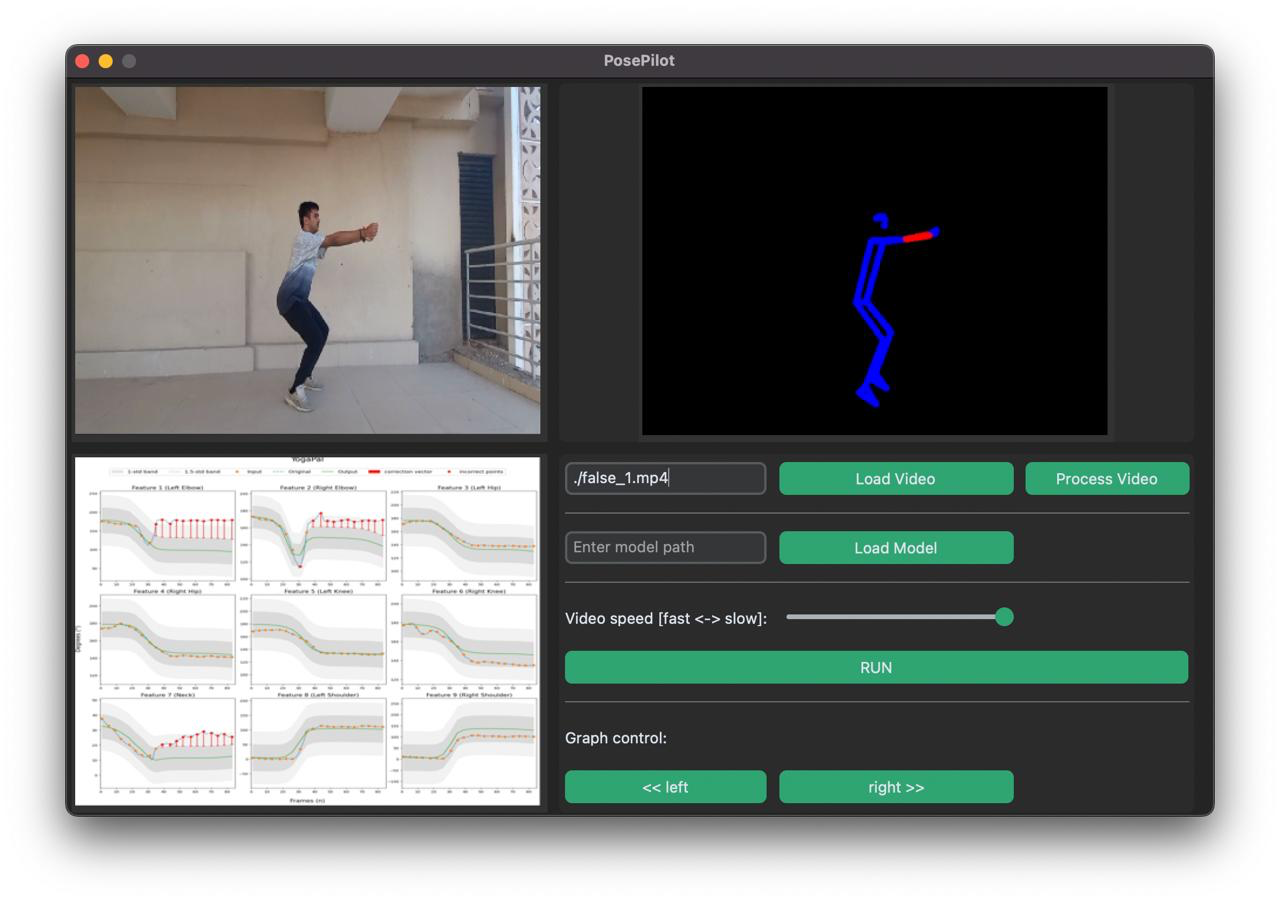}
    \caption{Pose Recognition and Personalized Corrective Feedback GUI}
    \label{fig: GUI}
\end{figure}

The performance of the model was evaluated using a forecasting task on the in-house test dataset, where it achieved an average mean square error (MSE) of 0.00138. The MSE values for different \textit{Asanas} are listed in Table \ref{tab:classification_report}. This low MSE indicates that the model was highly accurate in predicting correct pose trajectories. As no external benchmarks are available (as detailed in Section \ref{sec:inhouse-dataset}), we therefore relied solely on our test dataset for evaluation.
Further, to validate the hypothesis that our model effectively captures and utilizes temporal patterns for pose correction, we conducted a series of experiments using deliberately incorrect \textit{asanas}. These tests were designed to assess the model’s ability to detect deviations from correct poses and provide feedback for correction. An example of this process is shown in Figure \ref{fig: Correction Graph}. After processing the video, all posture points that deviated more than 1.5 standard deviations from predicted pose were flagged as errors requiring correction. For each flagged instance, the system identified the specific body angles that needed adjustment and the direction of correction required. The goal was to bring the angles within the acceptable range, defined as within one standard deviation from the predicted value. This personalized feedback was presented to the user through a graphical user interface (GUI) (Figure \ref{fig: GUI}), which displayed a side-by-side comparison of the camera feed and the synchronized pose detection, highlighting the segments with posture errors. Additionally, the correction graph is also shown in GUI to support further analysis by the user. Overall, the system effectively detects misalignments and deviations in incorrect poses, delivering precise and personalized feedback to guide users toward correct posture and proper alignment.

\subsection{Edge Deployment}
To evaluate the performance of our pose recognition and pose correction models in real-world scenarios, we deployed them on a Raspberry Pi 4 with 4GB RAM and 64GB storage. The TensorFlow Lite framework was used to optimize model inference for the constrained hardware. By applying quantization to reduce model precision from FP32 to INT8, inference speed was further improved, with only an accuracy loss of only 1.0\% for pose recognition and 1.3\% for pose correction. Though feature extraction on video frames needs further optimization, the model itself had an inference speed of 330.65 FPS for pose recognition and 6.42 FPS for pose correction. Further evaluation was performed under varying environmental conditions, including changes in lighting and background complexity. The accuracy of pose recognition remained stable, with a variance of less than 2.5\%. The pose correction model also demonstrated consistent feedback accuracy, with an MSE fluctuation within 0.0006 of the baseline. These results confirm that PosePilot can run effectively on Raspberry Pi 4, balancing real-time performance and resource constraints for pose correction applications.

% need to improve conclusion
\section{CONCLUSION AND FUTURE WORK}
PosePilot successfully detects and corrects posture errors from diverse camera viewpoints, ensuring accurate execution of Yoga poses in real-world environments. This enhances both the safety and effectiveness of the practice. PosePilot not only achieves state-of-the-art performance for pose recognition but also provides a promising approach for pose correction with personalized feedback. We also demonstrated how BiLSTM could be both lightweight and efficient in such computer vision applications. Future work will focus on further optimizing the latency of our end-to-end pipeline, encompassing feature extraction and model inference, to achieve better real-time performance on an edge device. We also plan to evaluate PosePilot's effectiveness on end-to-end deployment for specific applications beyond Yoga in physiotherapy and sports coaching. We also aim to develop more explainable methods for pose correction to improve transparency and trust in provided corrections.

\section*{Acknowledgments}
The authors are thankful to Harish and Bina Shah School of AI \& CS at Plaksha University for providing the seed financial support for this research.

% Since pose correction is a regression task rather than classification, the use of LSTM networks is more suitable as they effectively capture temporal dependencies. In contrast, Transformer models, despite their accuracy in other domains, are not feasible for deployment on edge devices due to high computational and memory requirements.

\bibliographystyle{splncs04}
\bibliography{sample-main} 

\end{document}